\documentclass{article}
\usepackage[utf8]{inputenc}

\usepackage{graphicx}
\usepackage{subcaption}
\usepackage[section]{placeins}

\begin{document}

\title{Enhance Multimodal Transformer With External Label And In-Domain Pretrain: Hateful Meme Challenge Winning Solution}
\author{Ron Zhu \\
  \multicolumn{1}{p{.7\textwidth}}{\centering\emph{ALFRED SYSTEM}}}

\date{December 2020}

\maketitle
\begin{abstract}
Hateful meme detection is a new research area recently brought out that requires both visual, linguistic understanding of the meme and some background knowledge to performing well on the task.
This technical report summarises the first place solution of the Hateful Meme Detection  Challenge 2020, which extending state-of-the-art visual-linguistic transformers to tackle this problem. At the end of the report, we also point out the shortcomings and possible directions for improving the current methodology.
\end{abstract}

\section{Introduction}
The Hateful Memes Challenge\cite{kiela2020hateful} introduces the dataset designed to be truly multimodal by including confounders that prevent the model from exploiting unimodal prior. Despite the recent advances in multimodal reasoning, we only get 0.71 AUROC when applying state of the art multimodal models to the challenge, which is far behind from non-expert human performance. In this paper, we discuss the difference between the commonly seen multimodal reasoning task and the meme classification. And the approaches we build on top of the existing visual-linguistic models to improve the performance and achieving 0.845 AUROC on the hateful memes detection dataset.

\begin{figure}[!htb]
\centering
\includegraphics[scale=0.2]{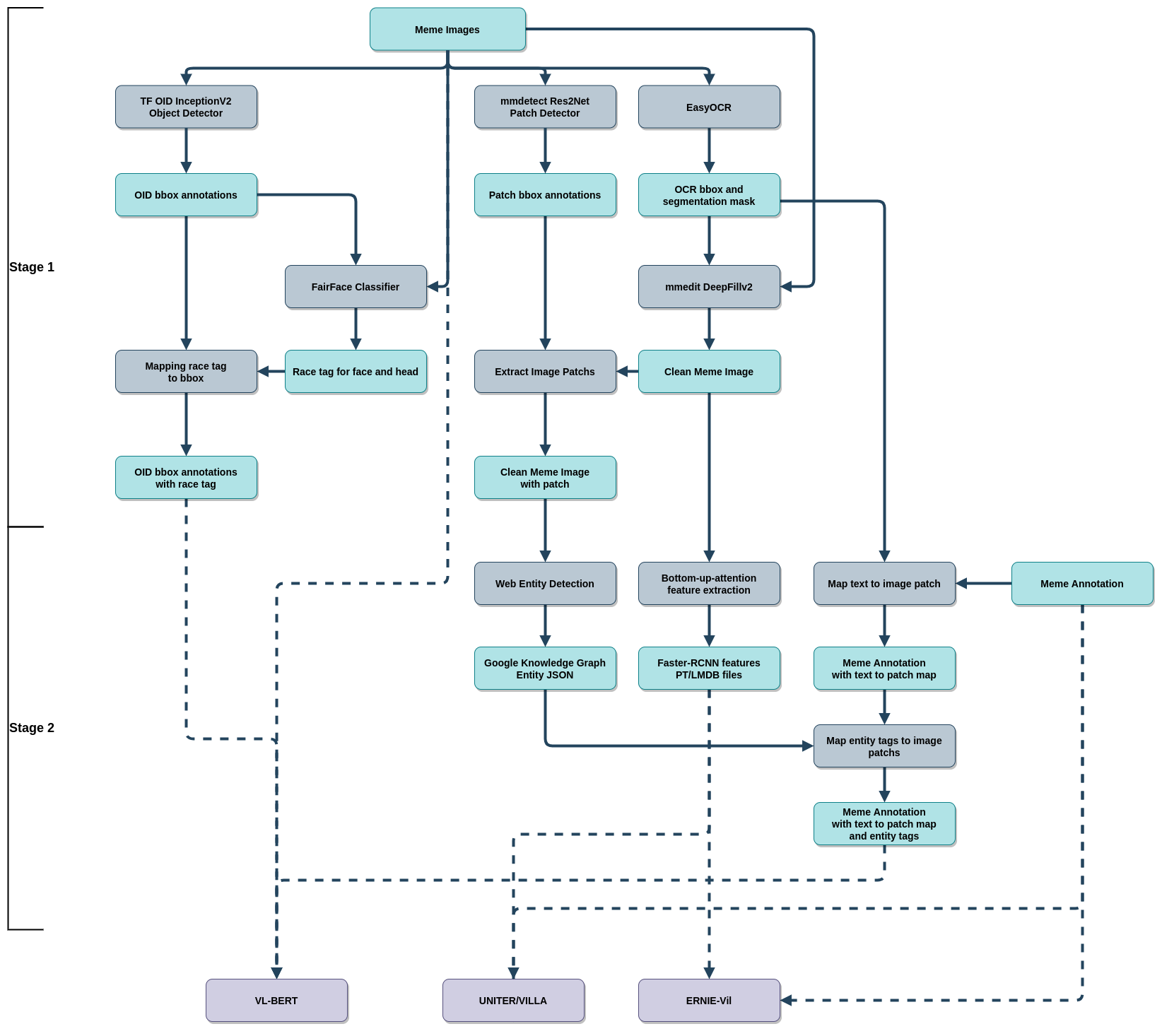}
\caption{Complete data processing pipeline}
\label{fig:meme_processing_flow}
\end{figure}

\FloatBarrier

\section{Problem Description}

In the typical visual-linguistic multimodal dataset, there is a direct relationship between the image and textual input. Hateful memes have their unique characteristic that is much different from the usual multimodal dataset. Compare to the multimodal dataset for a task like VCR\cite{zellers2019recognition} and VQA\cite{agrawal2016vqa}, only some of them have a direct relationship that appears in common multimodal down-stream tasks and pre-train datasets. Most of the time, the meme’s hateful part is linked by subtle wording or external knowledge about the real-world event. In that sense, hateful memes challenge is more of a visual-linguistic-knowledge multimodal classification problem.

Moreover, hateful memes are built on diverse topics and often contain novel objects that no off-the-shelf classifier will recognize, which is much different from the everyday object presented in typical multimodal datasets. Visual hints like injury, traditional costume wear by the subject, and historical scenes are hard to recognize and low on the sample number. Those factors make visual understanding part of the challenge tough to solve even without the lingual part.
The context of the visual modality context has an enormous impact on the meme’s polarity. Because visual hint is hard to recognize even for the state of the art image classifier or object detector, it becomes essential to incorporate information from a different source and format.

In some cases, the position of the meme caption will also affect the meaning of the meme. For example, intentionally place the caption of “wishing machine” on top of the female subject make the meme hateful, but if we place the caption on anything else instead of human will not. And this not commonly see and accounted for in the general visual-linguistic model.

\section{Approach}

The solution comprises four different VL transformer architecture ensemble, namely: VL-BERT\cite{su2020vlbert}, UNITER\cite{chen2020uniter}, VILLA\cite{gan2020largescale}, and ERNIE-Vil\cite{yu2020ernievil}. All the models besides ERNIE-Vil were modified for better performance of hateful meme detection. Furthermore, we conduct extra steps to extract more information from the hateful memes dataset for both training and inference.

\subsection{Background}
In the recent trend of applying the pre-trained transformer model to VL multimodal task have greatly improved the SOTA of multiple benchmarks. State of the art VL transformers like Oscar\cite{li2020oscar}, VILLA, ERNIE-Vil are utilize large scale dataset with paired image and text for pretraining. Although they achieve enormous improvement on the common sense VL tasks, pretrained VL transformer still performs poorly on hateful meme detection. The task requires the model to have a deep understanding of image, text, real-world events, and the ability to recognize inter-modal interaction between three modalities.

\subsection{Additional Data Source}
As forward-mentioned, we will need more than the image and text description of memes to do well on hateful meme detection. So at the first step of the data preprocessing, we will try to extract a few critical information from the data source using Web Entity Detection and FairFace\cite{karkkainen2019fairface} classifier.

Memes often contain a reference to famous historical events, news, celebrities. And memes reference to news topic and different sub-culture trend are changing rapidly. It not uncommon to see new memes popping up every few days or weeks. So using static knowledge sources(ex: Wikipedia) alone may not be adequate for this problem. To deal with this challenge, we use Google Vision Web Entity Detection to capture the image’s context. Despite this, without taking the entity’s background knowledge and the relation between entities into account, we only solve half of the problem.

According to the Hateful Meme Challenge paper, amount all types of hateful memes, hateful meme related to race or ethnicity is the most prominent. Almost half of the hateful memes can be categorized to it. Furthermore, it is unrealistic to expect a pretrained VL transformer to learn to identify race with just a few thousand samples from the hateful memes dataset. Thus the inclusion of race and gender labels created from the FairFace classifier. We apply the classifier by first detecting and classifying every face or head in the image, then mapping the label back to the person’s bounding box with the largest overlapped area with the face. In the end, we should get the race and gender of most of the person to appear in the image unless the person’s head is obscured.

\subsection{Extend VL-BERT Visual-Linguistic Framework}

To jointly train multiple types of labels with the original meme image and text. We need a way to combine information from different sources and formats seamlessly. Also, the ability to explicitly link images and text to the external labels will help deal with the unique characteristic of memes and remove the need for learning how to use external labels from scratch.

Inspire by Oscar, we represented all external labels as a special type of text token and linked it to the general area in the image or a special image region using visual feature embedding. By doing so, we also create the implicit link between text tokens that have related image features. For example, text token “bowler hat” with the image region of a hat and “man dressed in a three-piece suit” with the image region covers the entire person wearing the hat/suit. With a versertail framework like this, we can easily add more types of labels to handle different types of memes in the future.

\begin{figure}[!htb]
\centering
\includegraphics[scale=0.22]{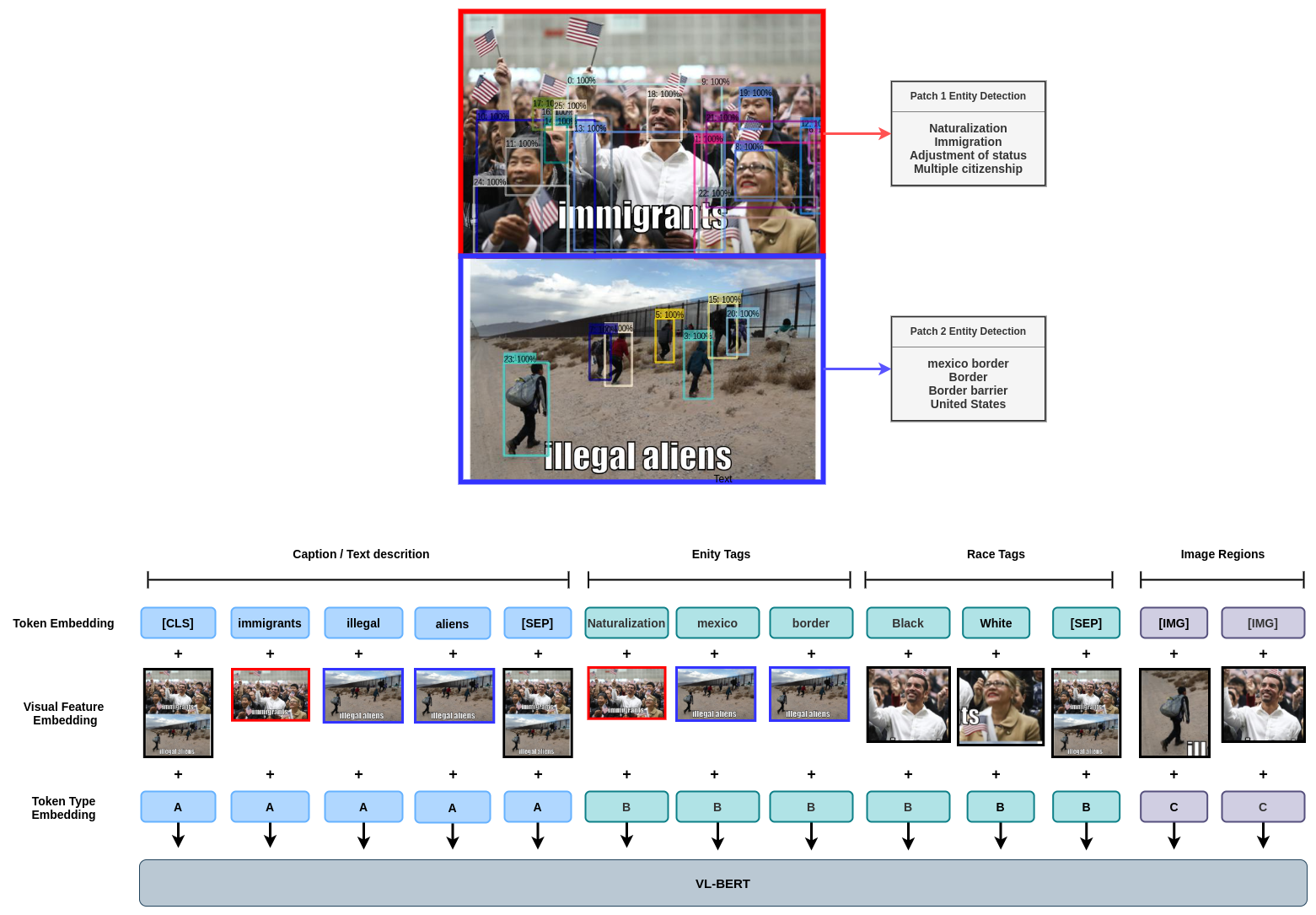}
\caption{Architecture of extened VL-BERT}
\label{fig:universe}
\end{figure}

\FloatBarrier

\subsection{UNITER-ITM}
Many pretrained VL transformer inspires by BERT\cite{devlin2019bert} also include the VL adaptation of the NSP(next sentence prediction) task - ITM(image text matching). Although some research shows that the inclusion of ITM tasks gives no benefit or even reduces the performance on the downstream task, we can still find it helpful depending on the task on hand. 
In the case of hateful memes detection, there is an apparent relationship between it and ITM; namely, the meme with the description(text) aligning the image’s content is most probably benign, but meme may not necessarily being hateful when they are not aligning.
It may just the meme’s description and image being unrecognizable or hard to understant by the model, or actually being hateful and comparing people to object or animals. 

With the UNITER pre-trained transformer, we keep the ITM head pre-trained on ITM and WRA(word region alignment) tasks instead of randomly initializing the classifier head weight. Then swap the weights of class 0 of classifier head with class 1, so now class “image-text-match” will become “benign” and “image-text-not-match” become “hateful”.

Suggest by recent research\cite{cao2020scene} both single-stream or dual-stream VL transformers are biasing toward linguistic modal.  By reusing ITM head as the starting point, we should get an extra benefit of reducing linguistic bias due to the face that ITM and WRA require both visual and linguistic feature to work, which is not always true for another pre-train task like MRC(masked region classification) or MLM(masked language modeling).

\begin{figure}[!htb]
\centering
\includegraphics[scale=0.5]{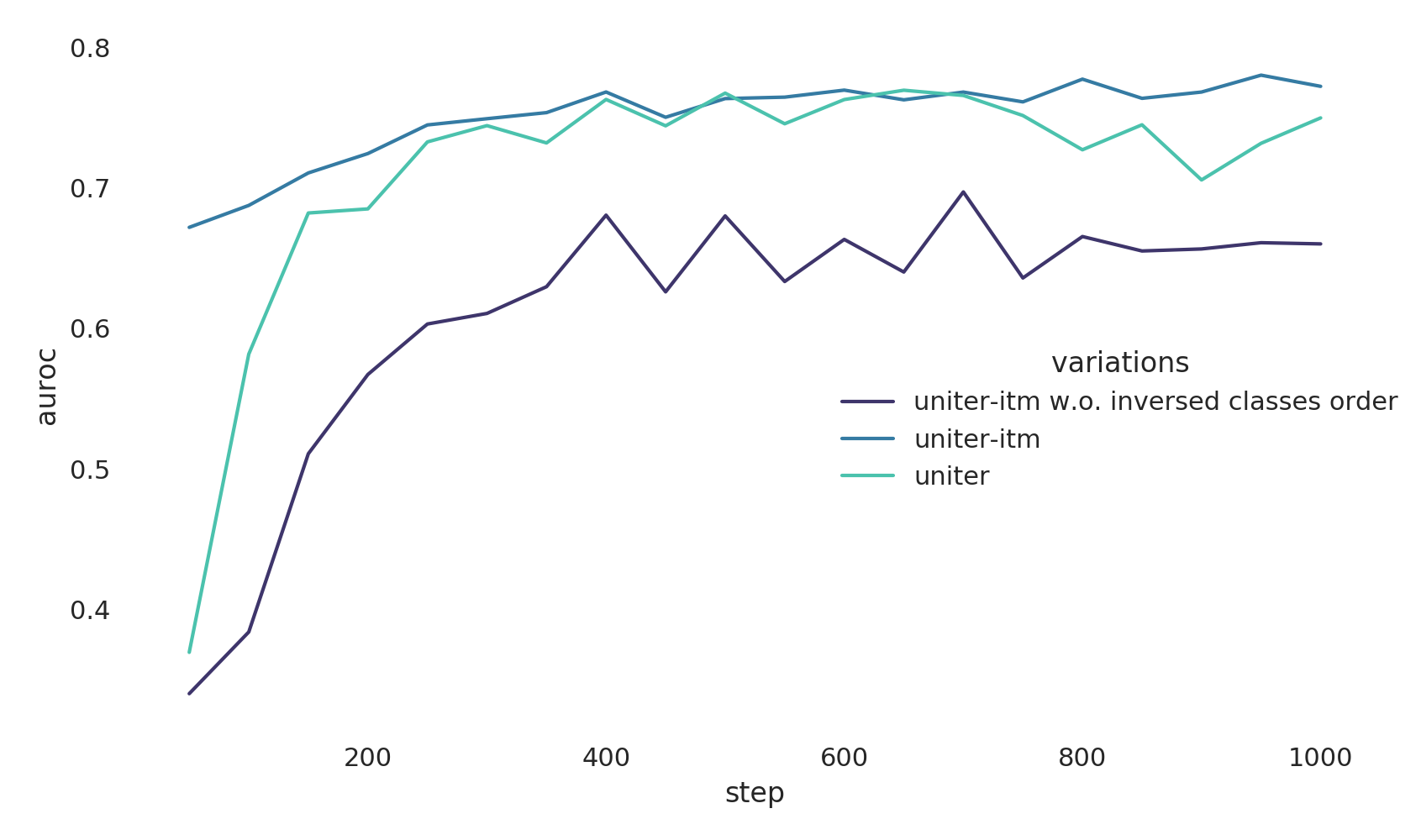}
\caption{UNITER-ITM}
\label{fig:UNITER-ITM}
\end{figure}

\FloatBarrier

\subsection{ERNIE-Vil}
ERNIE-Vil is the VL model pretrained on information extracted from the scene graph to teach the model about fine-grained object attribute and relationship model. Being the STOA model on the VCR leaderboard, ERNIE-Vil can achieve competitive performance to extended VL-BERT and UNITER-ITM with some hyper-parameter tuning. Including ERNIE-Vil into the final ensemble add diversity to the prediction and improve generalization.

\section{Experiment}
\subsection{Apply entity/race/gender tags}
By applying the new labels obtained from the external model to extended VL-BERT, our accuracy get a significant boost on the dev-seen set of hateful memes dataset. Nevertheless, when we apply the same entity tags to other single stream models like UNITER and Oscar, they give worse or almost no performance difference. Then apply the same trick on ERNIE-Vil give a noticeable jump in performance. While the difference in architecture and training target is small between single-stream models we tested, entity tags’ effect differs. There are a few potential reasons that may cause this phenomenon: 

1) Lack of explicitly visual-linguistic fusion mechanism (VL-BERT’s visual embedding for every token, and ERNIE-Vil’s cross-modality layers) make it hard to learn how to utilize a new type of input format with a small dataset like the one we are using.

2) Lower lingual representation quality(VL-BERT has been pre-trained with text only MLM task, ERNIE-Vil have the independent text modality stream) let the entity tags that are represented in the text become less effective. 

\begin{figure}[!htb]
\centering
\includegraphics[scale=0.5]{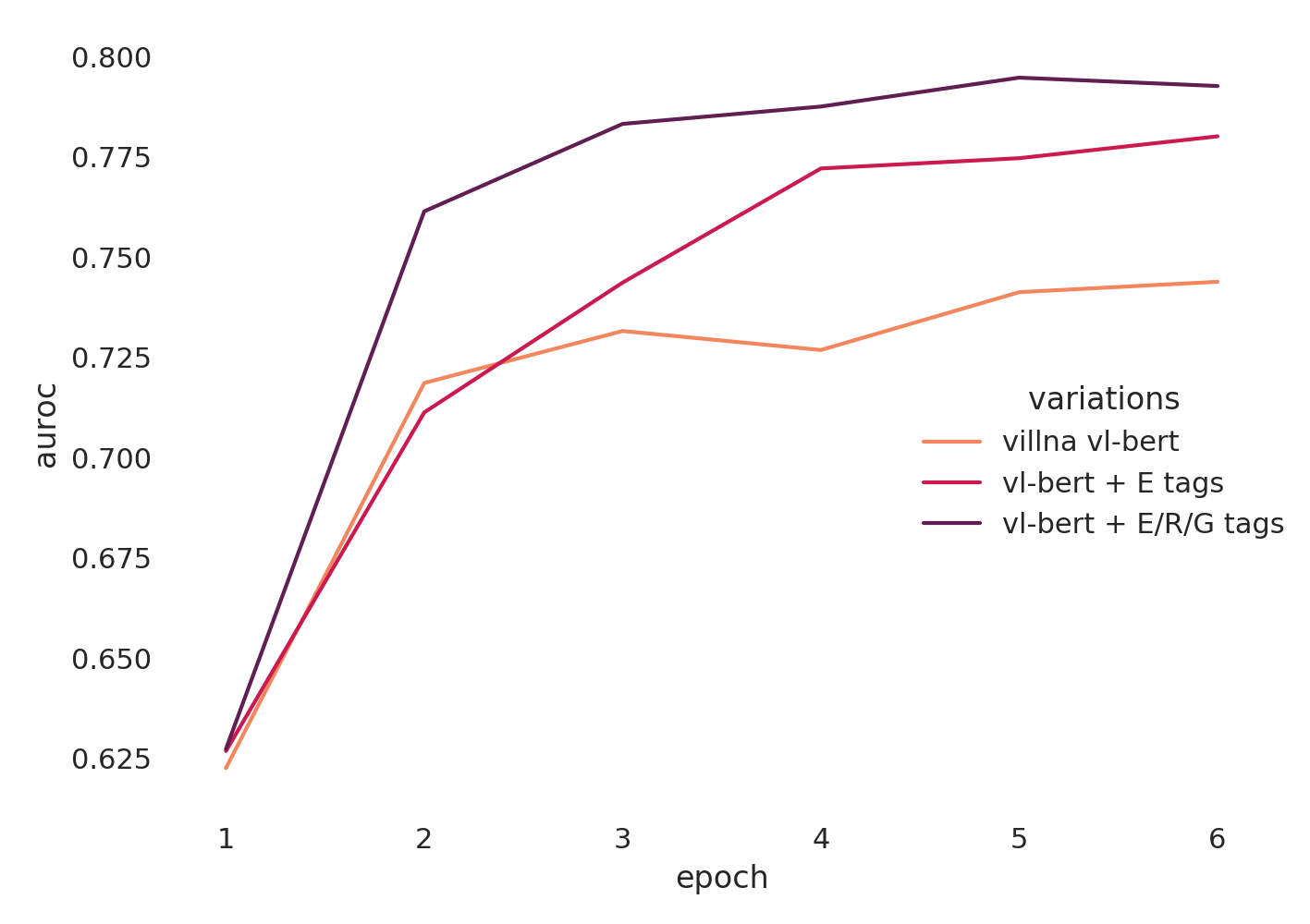}
\caption{dev-seen AUROC of extended VL-BERT with entity/race/gender tags}
\label{fig:vl-bert-val-auroc}
\end{figure}

\begin{figure}
\centering
\begin{subfigure}{.5\textwidth}
  \centering
  \includegraphics[width=.9\linewidth]{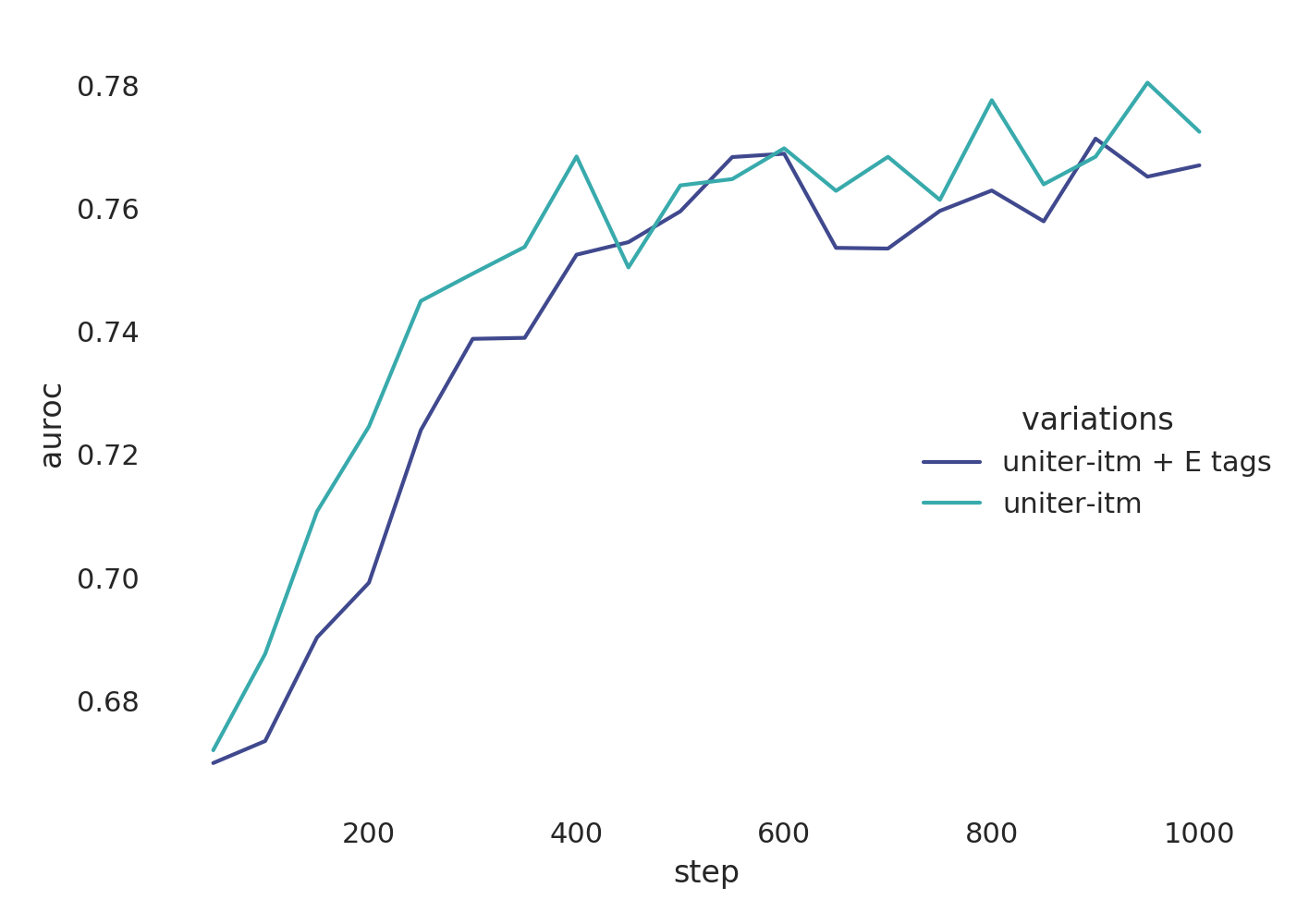}
  \caption{UNITER-ITM with and w/o entity tags}
  \label{fig:uniter-entity}
\end{subfigure}%
\begin{subfigure}{.5\textwidth}
  \centering
  \includegraphics[width=.9\linewidth]{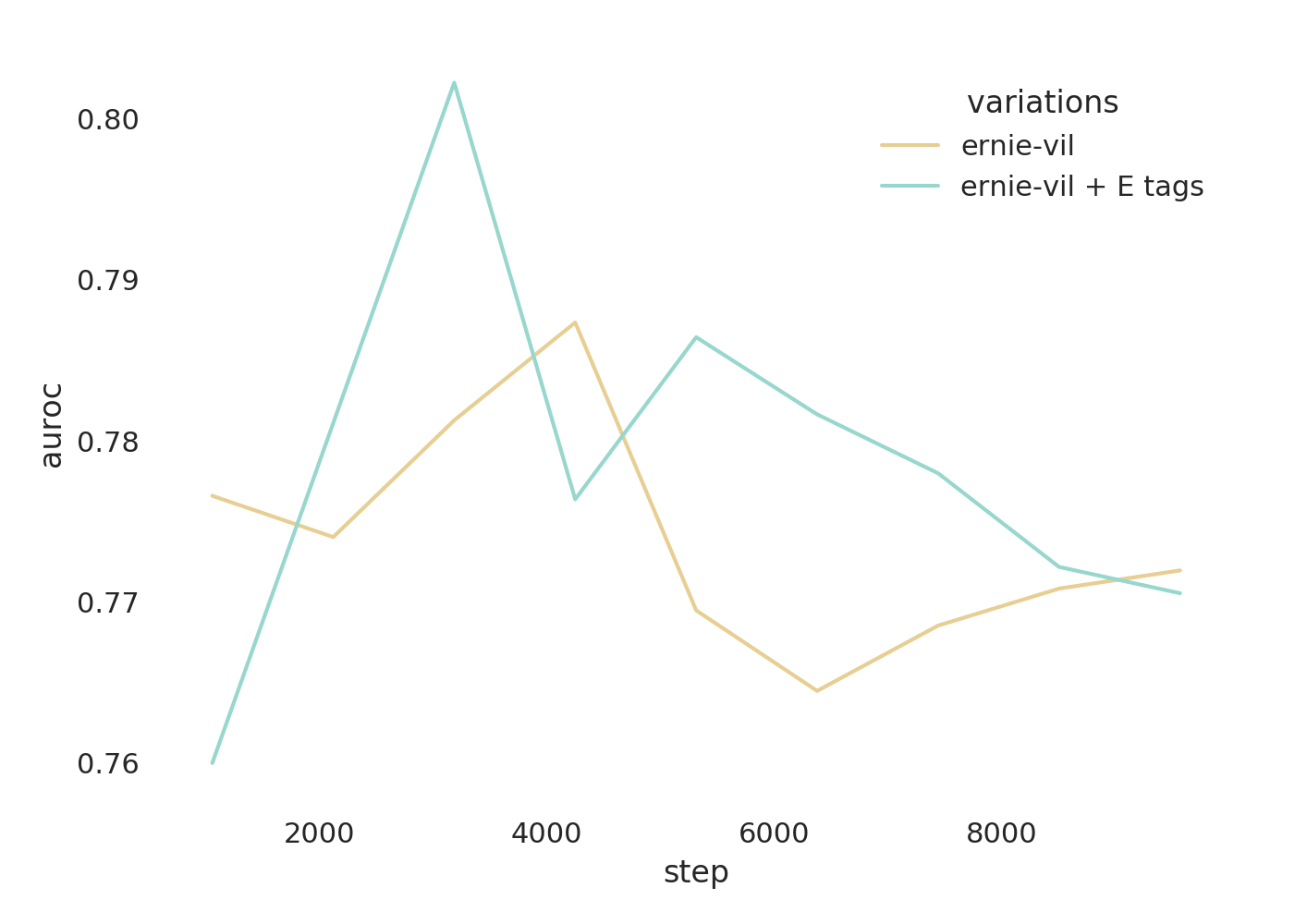}
  \caption{ERNIE-Vil with and w/o entity tags}
  \label{fig:ernie-entity-clean}
\end{subfigure}
\caption{Hateful Memes dev-seen set AUROC}
\label{fig:test}
\end{figure}

\FloatBarrier

\subsection{Reuse of UNITER ITM head}
As shown in the figure.6, UNITER-ITM compare to randomly initialized classification head: 

1) seen not as easy to overfit. 

2) getting better performance. (0.778 vs. 0.765 AUROC). 

3) may suffer less bias toward text modality, thanks to ITM and WRA pre-train task require both modal to work.

The high accuracy of UNITER-ITM at the very start of the training may also indicate a correlation between ITM pre-train task and hateful memes classification. This leaves an interesting question of if there is another pre-train task for the VL model that also has the same high correlation to hateful memes classification.

\begin{figure}[!htb]
\centering
\includegraphics[scale=0.5]{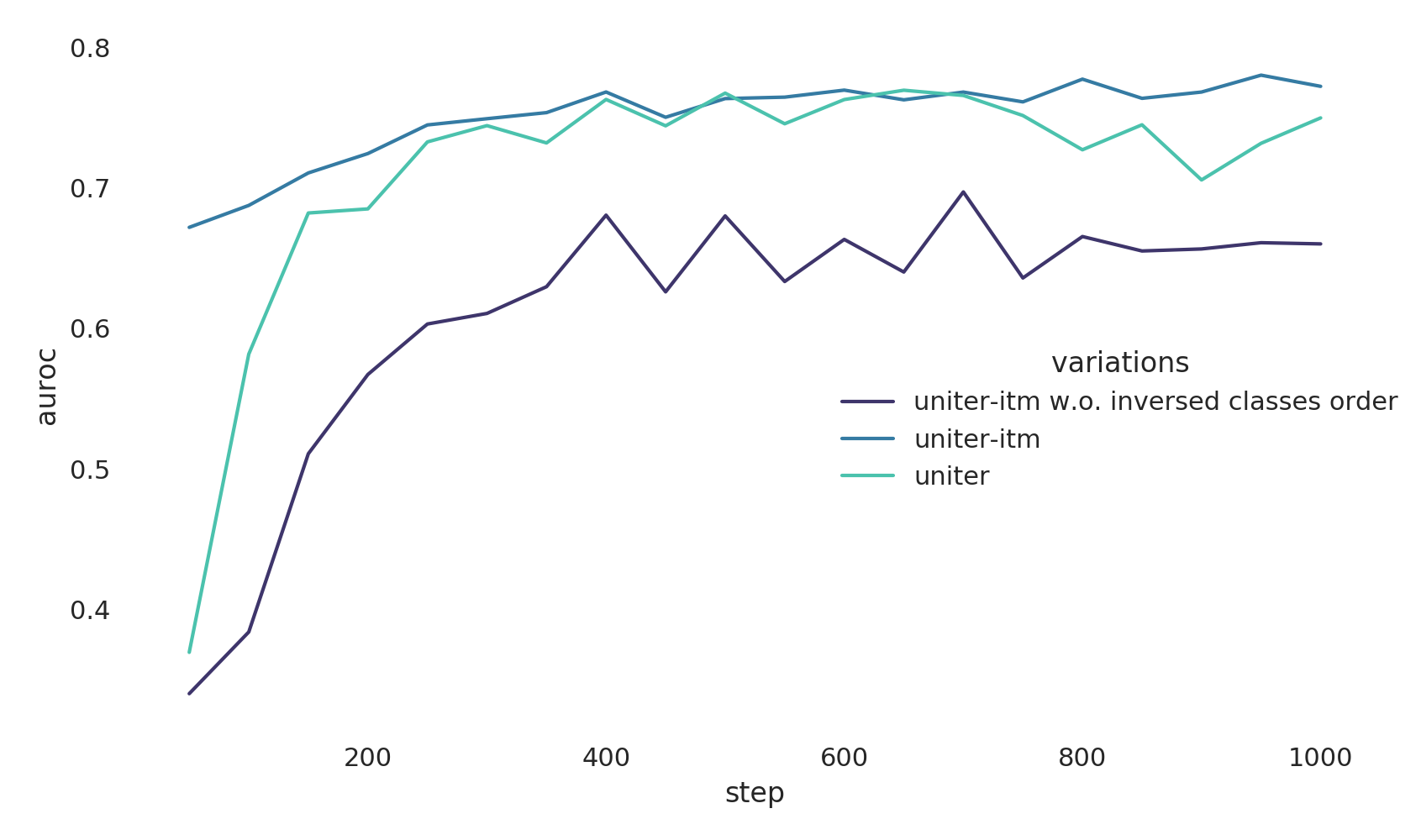}
\caption{dev-seen AUROC of UNITER w/ and w/o reuse ITM head}
\label{fig:uniter-itm}
\end{figure}

\section{Feature Direction}
Hateful meme detection is currently one of the few tasks that require the model to identify entities and relationships from image, text, and external knowledge base. In the Hateful Memes dataset, hateful metaphor hidden between relation visual, linguistic, knowledge base entities are common but extremely hard to spot for the machine. Most of the pre-trained models currently apply to hateful meme detection are only take image and text into account, and the pre-train dataset only contained everyday object and simple description that gives a shallow explanation of the image. Although the pre-trained language model used in some VL models can be treated as a knowledge base and have decent performance in the task like open-domain QA. However, this kind of implicit knowledge base is hard to update and apply to a different domain. Several research attempts to resolve this problem by letting the model learn to use the external knowledge base instead of remembering everything inside the model’s parameters, for example, RAG\cite{lewis2020retrievalaugmented}. 

A better and more flexible source of knowledge that provides in-depth information about the entity that appears in the meme should help reduce the performance gap between machine and human. One of the promising directions also comes from the field of open-domain QA: knowledge graph enhanced pre-train transformers(language model). There are many ways we can inject external knowledge into pre-trained transformers like K-Adapter\cite{wang2020kadapter}, K-BERT\cite{liu2019kbert}. One of them, MHGRN\cite{feng2020scalable}, is especially interesting to me. Due to its plug-and-play nature, MHGRN can place after any pre-trained transformer; in our case, any one of the VL transformers will work. This gave us many options and flexibility on how we want to handle the visual-linguistic part of the problem. Moreover, the attention GNN used in MHGRN makes it easier to scale the model to a larger graph, which may be a critical part of the problem when dealing with the subtle relationship between key entities inside a meme. Despite the advantage mentioned above, using a knowledge-graph-based method to solve hateful meme detection still come with a few challenges. To cover the diverse topic meme is referencing from, we will also need a sizeable knowledge graph like wiki-data or ConceptNet\cite{speer2018conceptnet} to compensate for it. For every meme we are dealing with, we need to extract a subgraph that contains the entity that appears in the meme’s image or text and an entity that is potentially related to the context of the meme. Entities in a large scale knowledge like wiki-data usually have dozens of type of relation per entity. Thus building a subgraph that both includes enough information and has a reasonable size will be challenging.

\section{Conclusion}
In this paper, we propose two methods to improve pre-trained visual-linguistic transformers’ performance on the task of hateful meme detection. The first approach focuses on building a visual-linguistic framework on top of VL-BERT to incorporate pieces of information that are almost impossible to learn from the hateful meme challenge dataset directly. The second approach tries to utilize the entire pre-trained model to get a better score on meme classification. In spite of the improvement brought by those two approaches, there is still much space to improve. Furthermore, we also point out some of the promising research directions that may give the model a better understanding of the memes. 

\bibliographystyle{plain}

\begin{thebibliography}{10}

\bibitem{agrawal2016vqa}
Aishwarya Agrawal, Jiasen Lu, Stanislaw Antol, Margaret Mitchell, C.~Lawrence
  Zitnick, Dhruv Batra, and Devi Parikh.
\newblock Vqa: Visual question answering, 2016.

\bibitem{cao2020scene}
Jize Cao, Zhe Gan, Yu~Cheng, Licheng Yu, Yen-Chun Chen, and Jingjing Liu.
\newblock Behind the scene: Revealing the secrets of pre-trained
  vision-and-language models, 2020.

\bibitem{chen2020uniter}
Yen-Chun Chen, Linjie Li, Licheng Yu, Ahmed~El Kholy, Faisal Ahmed, Zhe Gan,
  Yu~Cheng, and Jingjing Liu.
\newblock Uniter: Universal image-text representation learning.
\newblock In {\em ECCV}, 2020.

\bibitem{devlin2019bert}
Jacob Devlin, Ming-Wei Chang, Kenton Lee, and Kristina Toutanova.
\newblock Bert: Pre-training of deep bidirectional transformers for language
  understanding, 2019.

\bibitem{feng2020scalable}
Yanlin Feng, Xinyue Chen, Bill~Yuchen Lin, Peifeng Wang, Jun Yan, and Xiang
  Ren.
\newblock Scalable multi-hop relational reasoning for knowledge-aware question
  answering, 2020.

\bibitem{gan2020largescale}
Zhe Gan, Yen-Chun Chen, Linjie Li, Chen Zhu, Yu~Cheng, and Jingjing Liu.
\newblock Large-scale adversarial training for vision-and-language
  representation learning, 2020.

\bibitem{karkkainen2019fairface}
Kimmo K{"a}rkk{"a}inen and Jungseock Joo.
\newblock Fairface: Face attribute dataset for balanced race, gender, and age.
\newblock {\em arXiv preprint arXiv:1908.04913}, 2019.

\bibitem{kiela2020hateful}
Douwe Kiela, Hamed Firooz, Aravind Mohan, Vedanuj Goswami, Amanpreet Singh,
  Pratik Ringshia, and Davide Testuggine.
\newblock The hateful memes challenge: Detecting hate speech in multimodal
  memes, 2020.

\bibitem{lewis2020retrievalaugmented}
Patrick Lewis, Ethan Perez, Aleksandara Piktus, Fabio Petroni, Vladimir
  Karpukhin, Naman Goyal, Heinrich Küttler, Mike Lewis, Wen tau Yih, Tim
  Rocktäschel, Sebastian Riedel, and Douwe Kiela.
\newblock Retrieval-augmented generation for knowledge-intensive nlp tasks,
  2020.

\bibitem{li2020oscar}
Xiujun Li, Xi~Yin, Chunyuan Li, Pengchuan Zhang, Xiaowei Hu, Lei Zhang, Lijuan
  Wang, Houdong Hu, Li~Dong, Furu Wei, Yejin Choi, and Jianfeng Gao.
\newblock Oscar: Object-semantics aligned pre-training for vision-language
  tasks, 2020.

\bibitem{liu2019kbert}
Weijie Liu, Peng Zhou, Zhe Zhao, Zhiruo Wang, Qi~Ju, Haotang Deng, and Ping
  Wang.
\newblock K-bert: Enabling language representation with knowledge graph, 2019.

\bibitem{speer2018conceptnet}
Robyn Speer, Joshua Chin, and Catherine Havasi.
\newblock Conceptnet 5.5: An open multilingual graph of general knowledge,
  2018.

\bibitem{su2020vlbert}
Weijie Su, Xizhou Zhu, Yue Cao, Bin Li, Lewei Lu, Furu Wei, and Jifeng Dai.
\newblock Vl-bert: Pre-training of generic visual-linguistic representations,
  2020.

\bibitem{wang2020kadapter}
Ruize Wang, Duyu Tang, Nan Duan, Zhongyu Wei, Xuanjing Huang, Jianshu ji,
  Guihong Cao, Daxin Jiang, and Ming Zhou.
\newblock K-adapter: Infusing knowledge into pre-trained models with adapters,
  2020.

\bibitem{yu2020ernievil}
Fei Yu, Jiji Tang, Weichong Yin, Yu~Sun, Hao Tian, Hua Wu, and Haifeng Wang.
\newblock Ernie-vil: Knowledge enhanced vision-language representations through
  scene graph, 2020.

\bibitem{zellers2019recognition}
Rowan Zellers, Yonatan Bisk, Ali Farhadi, and Yejin Choi.
\newblock From recognition to cognition: Visual commonsense reasoning, 2019.

\end{thebibliography}

\end{document}